\documentclass[sigconf]{acmart}
\setcopyright{none}
\renewcommand\footnotetextcopyrightpermission[1]{}
\setcopyright{none}
\usepackage{booktabs}
\usepackage{multirow}
\usepackage{xcolor}
\usepackage{graphicx}
\usepackage[ruled,vlined]{algorithm2e}
\SetKw{Continue}{continue}
\SetKwInput{KwIn}{Input}
\SetKwInput{KwOut}{Output}
\DontPrintSemicolon

\begin{document}

\title{Ray-Aware Pointer Memory with Adaptive Updates \newline for Streaming 3D Reconstruction}

\author[1]{Feifei Li\textsuperscript{*}}
\affiliation{%
  \institution{The Chinese University of HongKong, Shenzhen}
  \city{Shenzhen}
  \country{China}}
\email{feifeili1@link.cuhk.edu.cn}

\author[2]{Qi Song\textsuperscript{*}}
\affiliation{%
  \institution{Tsinghua University}
  \city{Beijing}
  \country{China}
}
\email{songqi@mail.tsinghua.edu.cn}

\author{Chi Zhang}
\affiliation{%
  \institution{The Chinese University of HongKong, Shenzhen}
  \city{Shenzhen}
  \country{China}
}
\email{chizhang1@link.cuhk.edu.cn}

\author{Rui Huang}
\affiliation{%
  \institution{The Chinese University of HongKong, Shenzhen}
  \city{Shenzhen}
  \country{China}
}
\email{ruihuang@cuhk.edu.cn}

\renewcommand{\shortauthors}{Li et al.}

\begin{abstract}
  Dense 3D reconstruction from continuous image streams requires both accurate geometric aggregation and stable long-term memory management. Recent feed-forward reconstruction frameworks integrate observations through persistent memory representations, yet most rely primarily on appearance-based similarity when updating memory. Such appearance-driven integration often leads to redundant accumulation of observations and unstable geometry when viewpoint changes occur.
  In this work, we propose a ray-aware pointer memory for streaming 3D reconstruction that explicitly models both spatial location and viewing direction within a unified memory representation. Each memory pointer stores its 3D position, associated ray direction, and feature embedding, allowing the system to reason jointly about geometric proximity and viewpoint consistency. Based on this representation, we introduce an adaptive pointer update strategy that replaces traditional fusion-based memory compression with a retain-or-replace mechanism. Instead of averaging nearby observations, the system selectively retains informative pointers while discarding redundant ones, preserving distinctive geometric structures while maintaining bounded memory growth.
  Furthermore, the joint reasoning over spatial distance and ray-direction discrepancy enables the system to distinguish between local redundancy, novel observations, and potential loop revisits in a unified manner. When loop candidates are detected, pose refinement is triggered to enforce global geometric consistency across the reconstruction.
  Extensive experiments demonstrate that the proposed ray-aware memory design significantly improves long-term reconstruction stability and camera pose accuracy while maintaining efficient streaming inference. Our approach provides a principled framework for scalable and drift-resistant online 3D reconstruction from image streams.
\end{abstract}



\begin{teaserfigure}
\begin{center}
   \includegraphics[width=\textwidth]{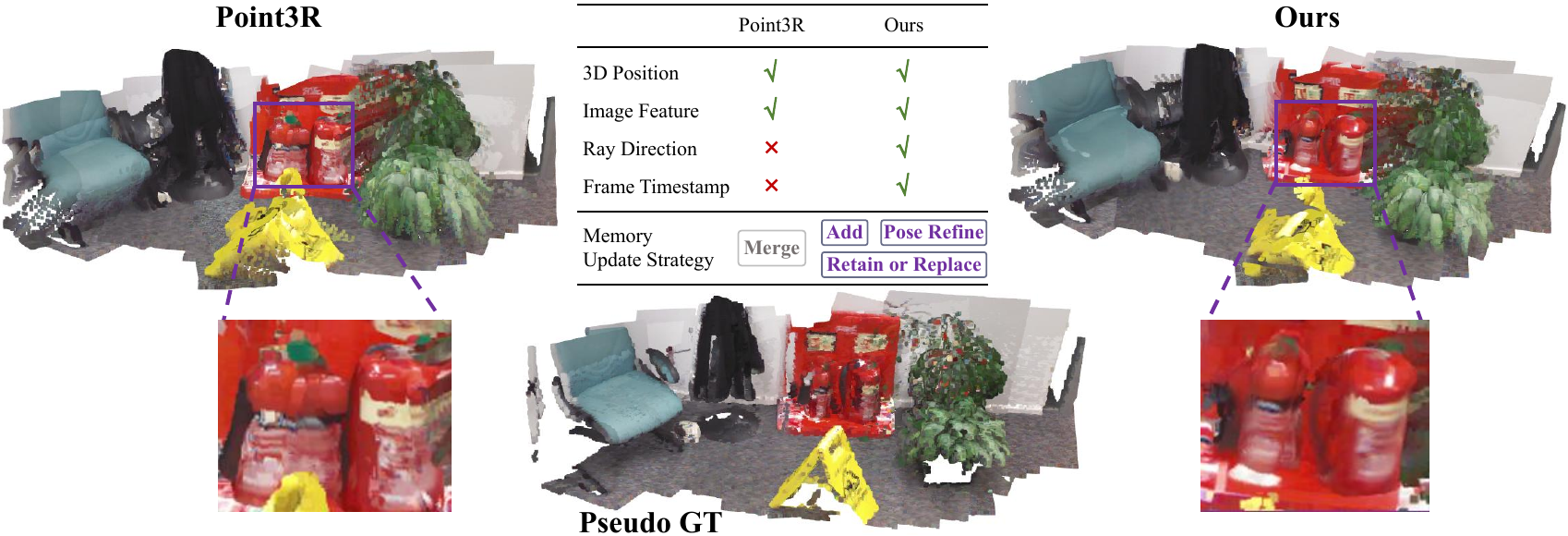}
\end{center}
   \caption{Comparison of visualized results of Point3R, our proposed method, and Pseudo GT. Pseudo GT of dense 3D model is obtained using the KinectFusion system. Compared with Point3R, our memory cache additionally stores ray direction and frame timestamp information. For the cache update strategy, we adopt a retain-or-replace scheme, and further perform pose refinement for detected loop closures. Our method outperforms Point3R in terms of geometric consistency.}
\label{fig:cover}
\end{teaserfigure}


\maketitle
\insert\footins{\footnotesize\textsuperscript{*}Equal contribution.\par}

\section{Introduction}
Dense 3D reconstruction from image sequences is a fundamental capability in computer vision, enabling applications in robotics, autonomous driving~\cite{zhang2025improving}, augmented reality, and digital scene modeling. 
Recently, feed-forward reconstruction frameworks have begun to replace the traditional geometric processing approaches~\cite{agarwal2011building-rome,schonberger2016sfm-revisited, wu2013towards,sweeney2015optimizing} with unified neural architectures. In particular, methods such as DUSt3R~\cite{wang2024dust3r} demonstrate that dense scene geometry can be recovered directly from image pairs by predicting pointmaps within a shared coordinate system. This paradigm significantly simplifies the reconstruction process by bypassing explicit triangulation and bundle adjustment. Nevertheless, while effective for pairwise matching, extending these methods to long image streams introduces significant new challenges.

Two main paradigms have been explored for long-sequence reconstruction. The first jointly processes multiple frames using global attention mechanisms, as seen in VGGT~\cite{wang2025vggt}, Spann3R~\cite{wang2025Spann3r}, and Fast3R~\cite{yang2025fast3r}. Although these approaches achieve strong global consistency, their computational complexity scales quadratically in long-horizon streaming scenarios.
In contrast, the second paradigm improves scalability by incrementally updating the scene representation as new observations arrive, enabling online reconstruction~\cite{yuan2026infinitevggt,chen2025long3r,chen2025ttt3r,deng2025vggt-long,lan2025stream3r,maggio2025vggt-slam,maggio2026vggt-slam2.0,wu2025point3r}. For instance, CUT3R~\cite{wang2025cut3r} and Point3R~\cite{wu2025point3r} maintain feature tokens or spatial pointers as scene memory elements.
These elements act as historical anchors representing the reconstructed 3D geometry, allowing incoming frames to be continuously matched and fused into the global scene via feature similarity or attention mechanisms.

However, avoiding incorrect associations between new observations and the stored memory elements remains a critical bottleneck for methods like Point3R, despite their use of both geometric positions and visual features. As viewpoints shift, geometrically adjacent regions may exhibit novel appearances or become severely occluded. Meanwhile, accumulated errors in predicted positions further complicate reliable matching. Consequently, relying solely on geometric positions and feature similarity may lead to cascading errors and eventual tracking drift, as visualized in Figure \ref{fig:cover}.

To address this issue, we propose a Ray-aware Pointer Memory for streaming 3D reconstruction. In contrast to existing memory representations, each pointer in our memory explicitly encodes not only its 3D position and visual features, but also the viewing direction of the observation ray and the source frame's timestamp (as shown in the middle part of Figure \ref{fig:cover}).
By incorporating the observation ray direction into the memory representation, the system gains crucial geometric context for disambiguating repeated observations under varying viewpoints.

Building on this representation, we introduce a Unified Observation Reasoner.
Observations that are spatially close and observed from similar viewpoints are interpreted as locally redundant measurements. In contrast, observations that are spatially close but observed from substantially different viewing directions are interpreted as potential loop revisits. Spatially distant observations correspond to previously unseen geometry and expand the reconstructed map. This unified reasoning framework enables the system to distinguish between redundancy, novelty, and loop closure in a geometrically meaningful manner.

To maintain a stable and compact memory representation, we further propose an Adaptive Pointer Update Strategy that replaces traditional fusion-based memory compression. Instead of averaging nearby observations—which may dilute distinctive geometric structures—we adopt a retain-or-replace policy that selectively preserves informative pointers while discarding redundant ones. This strategy maintains bounded memory growth while preserving the geometric discriminability of landmark observations.

By combining ray-aware pointer representation, unified observation reasoner, and adaptive pointer updates, the proposed framework improves the robustness and stability of streaming reconstruction over long sequences. Experiments on multiple datasets demonstrate that our method achieves competitive performance in dense 3D reconstruction, depth estimation, and camera pose estimation while maintaining efficient online inference.
Our contributions can be summarized as follows:
\begin{itemize}
    \item We introduce a pointer-based memory representation that explicitly stores both spatial position and viewing direction, enabling viewpoint-aware geometric reasoning in streaming reconstruction.
    \item We formulate memory integration as a joint geo-visual verification problem that distinguishes local redundancy, novel geometry, and loop revisits.
    \item We propose a retain-or-replace memory update strategy that preserves informative geometric landmarks while preventing uncontrolled memory growth.
\end{itemize}

Together, these contributions provide a principled framework for scalable and drift-resistant streaming 3D reconstruction from long image sequences.

\section{Related Works}

\subsection{Feed-forward Dense Reconstruction}

Recent advances in neural reconstruction have explored replacing traditional multi-stage pipelines with unified feed-forward architectures. Classical approaches to dense 3D reconstruction rely on feature matching~\cite{dusmanu2019d2-net,lowe2004distinctive,rublee2011orb,wu2013towards,lindenberger2023lightglue}, triangulation, bundle adjustment~\cite{triggs1999bundle,agarwal2010bundle}, which can produce accurate geometry but are computationally expensive and sensitive to noise accumulation in long sequences.
Later methods, such as Multi-view Stereo (MVS)~\cite{fu2022geo-neus,schonberger2016pixelwise,wei2021nerfingmvs}, Neural Radiance Fields (NeRF)~\cite{mildenhall2021nerf}, and 3D Gaussian Splatting (3DGS)~\cite{kerbl20233dgs}, use known camera poses to reconstruct geometry or represent the scene with high quality.

More recent methods instead learn to directly infer scene geometry from images. In particular, DUSt3R~\cite{wang2024dust3r} demonstrates that dense geometry can be recovered by predicting pointmaps from image pairs within a shared coordinate system, significantly simplifying the reconstruction pipeline. Subsequent works further improve this paradigm. MASt3R~\cite{leroy2024mast3r} introduces stronger geometric grounding to improve metric consistency, while MonST3R~\cite{zhang2024monst3r} extends the approach to handle scenes with motion. These methods demonstrate that feed-forward models can effectively recover dense geometry without explicit triangulation or bundle adjustment.

However, most of these approaches operate on image pairs. When extended to multi-view or long-sequence settings, they typically require an additional global alignment stage to reconcile pairwise predictions. This post-hoc alignment introduces additional optimization steps and becomes increasingly expensive as the sequence length grows, limiting scalability in streaming scenarios.

\subsection{Memory-Based Streaming Reconstruction}

To address the scalability limitations of pairwise reconstruction, recent works have explored streaming reconstruction frameworks that maintain a persistent memory representation of previously observed scene information. Instead of jointly processing all frames, these methods incrementally update the scene representation as new observations arrive.

Memory-based approaches can be broadly categorized into implicit feature memory~\cite{wang2025cut3r,chen2025ttt3r} and explicit spatial memory~\cite{10610868, wu2025point3r}.

Implicit memory approaches store latent representations of previous frames and integrate new observations through attention mechanisms. For example, Spann3R~\cite{wang2025Spann3r} maintains a feature memory that stores encoded frame representations and updates them using cross-attention with incoming frames. CUT3R introduces a bounded token-based memory state that evolves over time, enabling efficient streaming inference. While these approaches improve computational efficiency, the fixed-capacity memory inevitably requires overwriting earlier information as new frames are processed.

Explicit spatial memory approaches instead anchor memory entries to geometric locations in the scene. Point3R~\cite{wu2025point3r} introduces a pointer-based spatial memory in which each memory entry corresponds to a 3D point in the global coordinate system. By associating memory entries with spatial positions, this representation provides stronger geometric structure and prevents uncontrolled feature accumulation. To maintain memory compactness, nearby pointers are typically merged using fusion-based strategies.

Despite their efficiency, existing memory-based reconstruction frameworks largely rely on appearance similarity to determine whether new observations correspond to existing memory entries. However, appearance similarity alone does not uniquely determine spatial identity. Repetitive textures may cause observations from different locations to appear visually similar, while viewpoint changes may significantly alter the appearance of the same physical point. As a result, appearance-driven memory updates may lead to ambiguous associations and gradual drift in long sequences.

\subsection{Loop Closure and Geometric Consistency}

Loop closure is a fundamental component of classical SLAM systems. When the camera revisits a previously observed region, loop detection introduces long-range geometric constraints that allow accumulated pose errors to be corrected through pose graph optimization. Classical visual SLAM systems typically detect loops through place recognition and verify them using geometric constraints~\cite{maggio2025vggt-slam,maggio2026vggt-slam2.0}.

In contrast, most recent feed-forward reconstruction frameworks do not explicitly incorporate loop reasoning mechanisms. In memory-based reconstruction models such as CUT3R and Point3R, repeated observations are typically treated as standard memory updates, and pose inconsistencies are implicitly absorbed into latent representations rather than being explicitly corrected. As a result, pose drift may accumulate when the camera revisits previously explored areas.

Our work builds upon the explicit spatial memory paradigm while introducing ray-aware geometric reasoning to address the ambiguity of appearance-based memory updates. By storing both spatial position and viewing direction within each memory pointer, our representation enables the system to reason jointly about geometric proximity and viewpoint consistency. This design allows the system to distinguish between local redundancy and loop revisits in a unified manner and enables loop-triggered pose refinement during streaming reconstruction.

\section{Methods}

\begin{figure*}[htp]
\begin{center}
   \includegraphics[width=\textwidth]{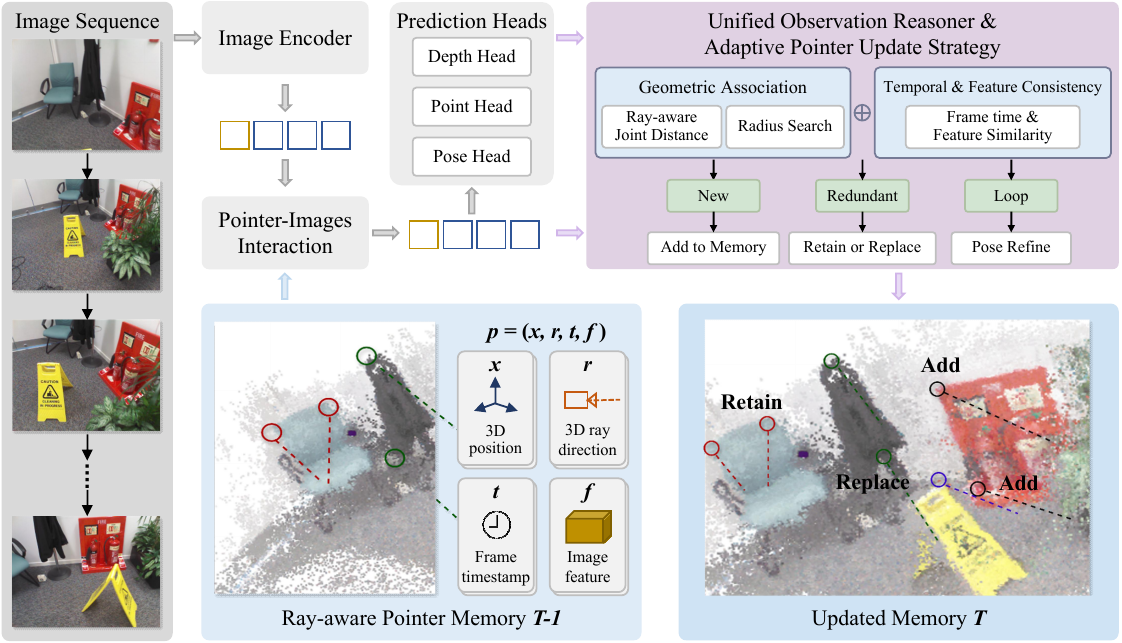}
\end{center}
   \caption{\textbf{Overview of the proposed ray-aware pointer-based streaming reconstruction pipeline.}
Each incoming frame is encoded into dense image features and interacts with a persistent 3D pointer memory through a pointer–image interaction module. Dedicated heads predict camera pose, depth, and point representations in a unified global coordinate system. Each pointer stores explicit spatial elements, including 3D position, viewing direction, frame timestamp, and learned features. The Memory Encoder module and Adaptive Points Update module incrementally refine the pointer set using pose-aware geometric consistency, selectively preserving informative pointers while filtering redundant observations, enabling stable and continuous streaming 3D reconstruction.}
\label{fig:pipeline}
\end{figure*}

We propose a ray-aware pointer memory framework for streaming 3D reconstruction. As shown in Figure~\ref{fig:pipeline}, the key idea is to explicitly model both spatial position and viewing direction for each memory entry, enabling the system to reason about geometric identity across changing viewpoints. Building on this representation, we introduce ray-guided geo-visual reasoning and an adaptive pointer update mechanism that enable stable integration of observations in long image streams.

\subsection{Problem Formulation}
We consider the problem of \textit{streaming 3D reconstruction} from a sequence of images:
\begin{equation}
    \mathcal{I} = \{I_1, I_2, \dots, I_T\}
\end{equation}
where each frame is captured from a camera with pose $P_t$.

The goal is to incrementally reconstruct a scene representation $\mathcal{M}$ while integrating observations from incoming frames.

A central challenge in streaming reconstruction is \textit{observation association}. When a new frame is processed, the system must determine whether newly observed pixels correspond to previously observed \textit{physical scene points} or represent new geometry.

Formally, for two observations $o_i$ and $o_j$, the system must determine:
\begin{equation}
    o_i \equiv o_j
\end{equation}
i.e., whether they correspond to the same physical point in the scene.

Existing memory-based reconstruction frameworks typically rely on \textit{appearance similarity} to determine observation identity. However, appearance similarity does not uniquely determine spatial identity. Two observations may have similar visual features while corresponding to different physical locations, or may appear dissimilar despite observing the same point from different viewpoints.

To address this ambiguity, we introduce a memory representation that jointly encodes \textit{spatial location and viewing direction}, enabling geometric reasoning for observation association.

\subsection{Ray-aware Pointer Memory}
We represent the scene using a set of \textit{spatial pointers}:
\begin{equation}
    \mathcal{M} = \{m_1, m_2, \dots, m_N\}
\end{equation}
Each pointer corresponds to an observation of a physical point in the scene and stores structured information:
\begin{equation}
    m_k = \{ \mathbf{x}_k, \mathbf{r}_k, \mathbf{f}_k, t_k \}
\end{equation}
where $\mathbf{x}_k \in \mathbb{R}^3$ denotes 3D position, $\mathbf{r}_k \in \mathbb{S}^2$ denotes unit ray direction from camera to point, $\mathbf{f}_k \in \mathbb{R}^d$ denotes feature embedding, $t_k$ denotes frame index when the pointer was created.

Unlike previous memory representations that store only spatial coordinates or latent tokens, the ray-aware pointer explicitly encodes the \textit{viewing geometry} associated with the observation.

This representation captures the relationship between a physical point and the viewpoint from which it was observed. In real-world scenes, the appearance of a point may vary significantly with viewing direction due to perspective distortion, occlusion, or lighting effects. By storing the observation ray, the system gains additional geometric context that helps disambiguate repeated observations across viewpoints.

During streaming reconstruction, each incoming frame interacts with the pointer memory through a pointer–image interaction module that predicts depth, camera pose, and candidate 3D points. These candidate points are converted into new pointers and integrated into the memory using the update mechanism described below.

\begin{figure*}[htp]
\begin{center}
   \includegraphics[width=1.\linewidth]{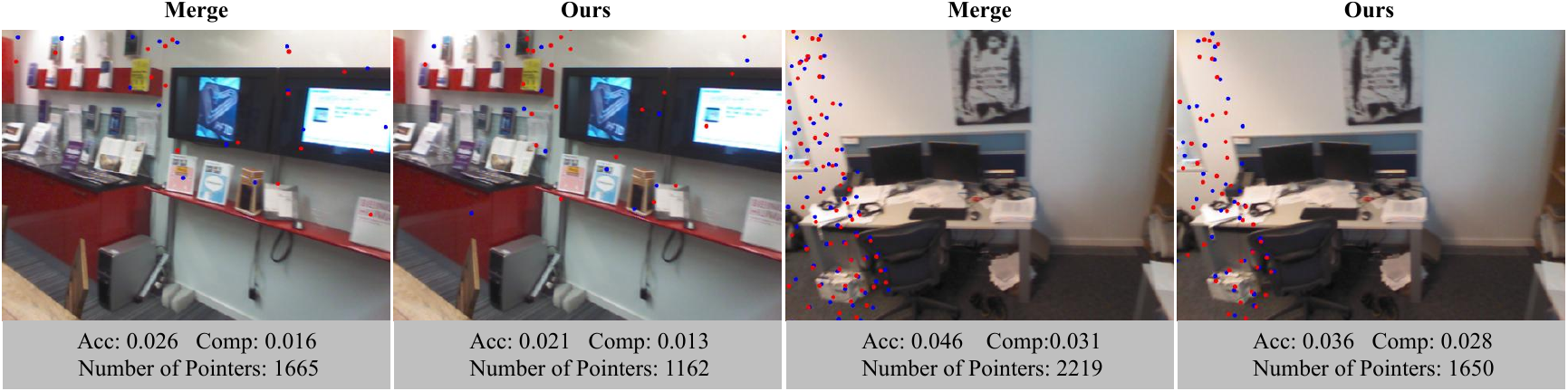}
\end{center}
   \caption{Illustration of the pointer update results for a given frame using the \texttt{merge} strategy and our \texttt{retain\_or\_replace} method. Blue pointers are replaced by red pointers. The metrics shown below the image represent the quantitative reconstruction performance of the current scene and the number of pointers stored in the memory cache after reconstruction. Our method significantly reduces the number of pointers stored in the memory cache while achieving improved reconstruction performance.
}
\label{fig:pointer_update}
\end{figure*}

\subsection{Unified Observation Reasoner}
When a new pointer $m_{new}$ is generated, the system must determine whether it corresponds to an existing memory pointer or represents new geometry.
To measure the similarity between pointers, we introduce a \textit{joint geometric distance metric} that considers both spatial proximity and viewing direction discrepancy.

Given a new pointer $m_{new}$ and an existing memory pointer $m_{k}$, we compute the following distances:

Spatial distance of the 3D positions of two points:
\begin{equation}
    d_{pos} = \| \mathbf{x}_{new} - \mathbf{x}_{k} \|_2
\end{equation}

Angular difference between observation rays that travel from camera to the points:
\begin{equation}
    d_{ang} = 1 - (\mathbf{r}_{new} \cdot \mathbf{r}_{k})
\end{equation}

The joint distance of spatial distance and angular difference is defined as:
\begin{equation}
    D(m_{new}, m_{k}) = \lambda_{pos} d_{pos} + \lambda_{ang} d_{ang}
\end{equation}
where $\lambda_{pos}$ and $\lambda_{ang}$ balance spatial proximity and viewpoint consistency.

This joint distance enables the system to interpret repeated observations in three distinct cases:
\begin{itemize}
    \item \textbf{Local redundancy.} When both spatial distance and angular difference are small, the observations correspond to the same physical point viewed from similar viewpoints.
    \item \textbf{Loop revisit.} When spatial distance is small but angular difference is large, the observations likely correspond to a revisited region observed from a different viewpoint.
    \item \textbf{Novel geometry.} When spatial distance is large, the observation corresponds to previously unseen geometry.
\end{itemize}

\subsection{Adaptive Pointer Update Strategy}

Traditional pointer-based reconstruction systems often use \textit{fusion-based updates}, in which nearby pointers are merged through feature averaging. While fusion reduces memory size, it may also dilute distinctive geometric structures and weaken the discriminability of landmark observations.

To preserve geometric information, we adopt a \textit{retain-or-replace strategy} instead of deterministic fusion.

For each new pointer $m_{new}$, we first search for nearby pointers within a spatial radius $R$:
\begin{equation}
    \mathcal{N}(m_{new}) = \{ m_k \in \mathcal{M} \mid \| \mathbf{x}_{new} - \mathbf{x}_{k} \|_2 < R \}
\end{equation}

If no neighboring pointer exists, the new pointer is added directly to memory.

If nearby pointers exist, we identify the closest pointer according to the joint distance metric. Instead of merging the pointers, we apply a stochastic retain-or-replace policy: either the new pointer or the existing pointer is discarded with equal probability.

As shown in Figure~\ref{fig:pointer_update}, this strategy prevents deterministic overwriting of earlier observations while preserving diverse geometric evidence across viewpoints. As a result, the memory maintains sharper geometric structures and avoids excessive averaging of landmark features. In our method, the total number of pointers in the memory cache is smaller, yet they are more uniformly distributed across the scene, resulting in stronger geometric representation capability.

\subsection{Loop Detection and Pose Refinement}
The ray-aware pointer representation also provides a natural mechanism for detecting loop revisits.

Two pointers $m_i$ and $m_j$ are considered loop candidates if:
\begin{equation}
\begin{alignedat}{3}
\| \mathbf{x}_i - \mathbf{x}_j \|_2 &<{}& \epsilon_{pos} &\quad& \text{(spatial)} \\
1 - (\mathbf{r}_i \cdot \mathbf{r}_j) &>{}& \epsilon_{ang} &\quad& \text{(angular)} \\
|t_i - t_j| &>{}& \Delta t &\quad& \text{(temporal)}
\end{alignedat}
\end{equation}
These conditions indicate that the observations correspond to spatially close locations observed from significantly different viewpoints after a long temporal interval.

When loop candidates are detected, we introduce pose constraints between the corresponding frames and perform pose graph refinement to reduce accumulated drift. After optimization, the pointer memory is updated under the refined coordinate system to maintain global consistency.
We also apply a Fisher information-based selection strategy~\cite{jiang2024fisherrf} to further refine and sparsify the pointers within the loop region. This procedure retains the most informative observations while discarding redundant ones, thereby improving the storage efficiency and structural stability of the memory cache.

\section{Experiments}

\begin{table*}[t]
\centering
\small
\resizebox{0.95\textwidth}{!}{
\begin{tabular}{lcc|cccccc|cccccc}
\toprule
& \multirow{3}{*}{Optim.} & \multirow{3}{*}{Onl.}
& \multicolumn{6}{c|}{\textbf{7-scenes}}
& \multicolumn{6}{c}{\textbf{NRGBD}} \\
\cmidrule(lr){4-9}\cmidrule(lr){10-15}
& & &
\multicolumn{2}{c}{Acc$\downarrow$} &
\multicolumn{2}{c}{Comp$\downarrow$} &
\multicolumn{2}{c|}{NC$\uparrow$} &
\multicolumn{2}{c}{Acc$\downarrow$} &
\multicolumn{2}{c}{Comp$\downarrow$} &
\multicolumn{2}{c}{NC$\uparrow$} \\
\cmidrule(lr){4-5}\cmidrule(lr){6-7}\cmidrule(lr){8-9}
\cmidrule(lr){10-11}\cmidrule(lr){12-13}\cmidrule(lr){14-15}
\textbf{Method} & & &
Mean & Med. & Mean & Med. & Mean & Med. &
Mean & Med. & Mean & Med. & Mean & Med. \\
\midrule
DUSt3R-GA~\cite{wang2024dust3r} & \checkmark &  &
\textbf{0.146} & \textbf{0.077} & \underline{0.181} & \textbf{0.067} & \textbf{0.736} & \textbf{0.839} &
\underline{0.144} & \textbf{0.019} & \underline{0.154} & \textbf{0.018} & \textbf{0.870} & \textbf{0.982} \\
MASt3R-GA~\cite{leroy2024mast3r} & \checkmark &  &
\underline{0.185} & \underline{0.081} & \textbf{0.180} & \underline{0.069} & \underline{0.701} & \underline{0.792} &
\textbf{0.085} & \underline{0.033} & \textbf{0.063} & \underline{0.028} & \underline{0.794} & \underline{0.928} \\
MonST3R-GA~\cite{zhang2024monst3r} & \checkmark &  &
0.248 & 0.185 & 0.266 & 0.167 & 0.672 & 0.759 &
0.272 & 0.114 & 0.287 & 0.110 & 0.758 & 0.843 \\
\midrule
Spann3R~\cite{wang2025Spann3r} &  & \checkmark &
0.298 & 0.226 & 0.205 & 0.112 & 0.650 & 0.730 &
0.416 & 0.323 & 0.417 & 0.285 & 0.684 & 0.789 \\
CUT3R~\cite{wang2025cut3r} &  & \checkmark &
0.126 & 0.047 & 0.154 & 0.031 & \underline{0.727} & \underline{0.834} &
0.099 & 0.031 & 0.076 & \underline{0.026} & \textbf{0.837} & \textbf{0.971} \\
Point3R~\cite{wu2025point3r} &  & \checkmark &
\underline{0.085} & \underline{0.046} & \underline{0.087} & \underline{0.030} & \textbf{0.739} & \textbf{0.854} &
\underline{0.077} & \textbf{0.030} & \underline{0.069} & 0.027 & \underline{0.835} & \textbf{0.971} \\
\textbf{Ours} &  & \checkmark &
\textbf{0.035} & \textbf{0.019} & \textbf{0.025} & \textbf{0.008} & {0.685} & {0.786} &
\textbf{0.061} & \underline{0.031} & \textbf{0.022} & \textbf{0.008} & 0.771 & \underline{0.926} \\
\bottomrule
\end{tabular}
}
\caption{Comparison of 7-scenes and NRGBD. Lower is better for Acc/Comp, higher is better for NC. We follow Point3R~\cite{wu2025point3r} to use “GA” to mark methods with global alignment, and use “Optim.” and “Onl.” to distinguish between optimization-based and online methods.}
\label{tab:mv_recon_7scenes_nrgbd}
\end{table*}

We evaluate the proposed ray-aware pointer memory framework on multiple tasks related to geometric perception, including \textit{dense 3D reconstruction}, \textit{monocular depth estimation}, and \textit{camera pose estimation}. These tasks jointly measure the geometric accuracy, spatial consistency, and pose stability of the reconstructed scene representation.

Our evaluation focuses on streaming scenarios where frames are processed sequentially and integrated into a persistent spatial memory.
\subsection{Experimental Setup}
\subsubsection*{Baselines}
We evaluate our method on three representative tasks: dense 3D reconstruction, monocular and video depth estimation, and camera pose estimation. As baselines, we compare against DUSt3R, MASt3R, MonST3R, Spann3R, CUT3R, and Point3R. Among these approaches, DUSt3R, MASt3R, and MonST3R operate on image pairs and therefore require an additional optimization-based global alignment stage when applied to streaming inputs. We follow the optimization-based global alignment (GA) stage conducted in Point3R~\cite{wu2025point3r}. We set $\lambda_{pos}=1$ and $\lambda_{ang}=0.1$ in all experiments.bundle

\begin{figure*}[t]
\begin{center}
   \includegraphics[width=0.95\linewidth]{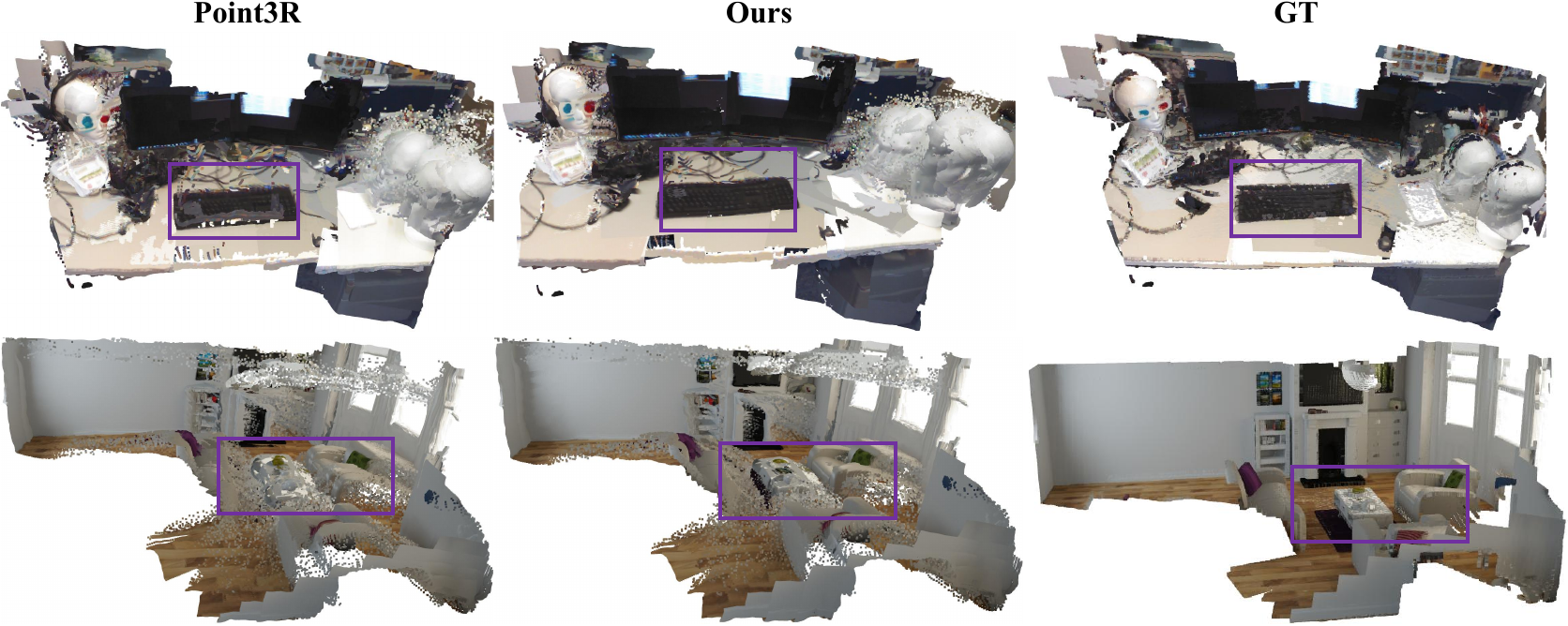}
\end{center}
   \caption{Visualized results of reconstruction on datasets NRGBD and 7scenes.}
\label{fig:visualized_recon}
\end{figure*}

\subsubsection*{Evaluation Tasks}
We evaluate the proposed method on three representative tasks:
Dense 3D reconstruction, Depth estimation, and Camera pose estimation.
These tasks collectively assess the ability of the system to maintain \textit{accurate geometry and stable pose estimation over time}.

\subsubsection*{Datasets}
For reconstruction evaluation, we use 7-Scenes and NRGBD. These datasets contain indoor scenes with challenging viewpoint changes and repeated structures.
Depth estimation is evaluated on NYU-v2~\cite{silberman2012nyu-v2}, Sintel~\cite{butler2012sintel}, Bonn~\cite{palazzolo2019bonn}, and KITTI~\cite{geiger2013kitti}.
Camera pose estimation is evaluated on ScanNet~\cite{dai2017scannet}, Sintel~\cite{butler2012sintel}, and TUM-dynamic~\cite{sturm2012tum}.
These datasets cover both \textit{static and dynamic scenes}, allowing us to analyze the robustness of the proposed method across diverse environments.

\begin{table*}[t]
\centering
\small
\resizebox{0.95\textwidth}{!}{
\begin{tabular}{l|cc|cc|cc|cc}
\toprule
& \multicolumn{2}{c|}{\textbf{NYU-v2 (Static)}} 
& \multicolumn{2}{c|}{\textbf{Sintel}} 
& \multicolumn{2}{c|}{\textbf{Bonn}} 
& \multicolumn{2}{c}{\textbf{KITTI}} \\
\cmidrule(lr){2-3}\cmidrule(lr){4-5}\cmidrule(lr){6-7}\cmidrule(lr){8-9}
\textbf{Method}
& Abs Rel$\downarrow$ & $\delta<1.25\,\uparrow$
& Abs Rel$\downarrow$ & $\delta<1.25\,\uparrow$
& Abs Rel$\downarrow$ & $\delta<1.25\,\uparrow$
& Abs Rel$\downarrow$ & $\delta<1.25\,\uparrow$ \\
\midrule
DUSt3R~\cite{wang2024dust3r} & 0.080 & 90.7 & 0.424 & 58.7 & 0.141 & 82.5 & 0.112 & 86.3 \\
MASt3R~\cite{leroy2024mast3r} & 0.129 & 84.9 & \textbf{0.340} & \textbf{60.4} & 0.142 & 82.0 & \underline{0.079} & \textbf{94.7} \\
MonST3R~\cite{zhang2024monst3r} & 0.102 & 88.0 & \underline{0.358} & 54.8 & 0.076 & 93.9 & 0.100 & 89.3 \\
Spann3R~\cite{wang2025Spann3r} & 0.122 & 84.9 & 0.470 & 53.9 & 0.118 & 85.9 & 0.128 & 84.6 \\
CUT3R~\cite{wang2025cut3r} & {0.086} & {90.9} & 0.428 & \underline{55.4} & {0.063} & \textbf{96.2} & 0.092 & 91.3 \\
Point3R~\cite{wu2025point3r}          & \underline{0.078} & \underline{92.3} & 0.395 & 55.0 & \underline{0.061} & {94.5} & {0.083} & \underline{94.6} \\
\textbf{Ours} & \textbf{0.073} & \textbf{93.1} & 0.376 & 56.6 & \textbf{0.059} & \underline{95.1} & \textbf{0.076} & {94.3} \\
\bottomrule
\end{tabular}
}
\caption{Depth evaluation on NYU-v2 (Static), Sintel, Bonn, and KITTI. Lower is better for Abs Rel; higher is better for $\delta<1.25$.}
\label{tab:depth_benchmarks}
\end{table*}

\begin{table*}[t]
\centering
\small
\resizebox{0.95\textwidth}{!}{
\begin{tabular}{lcc|ccc|ccc|ccc}
\toprule
\multicolumn{1}{c}{\textbf{Method}} &
\multicolumn{1}{c}{\textbf{Optim.}} &
\multicolumn{1}{c|}{\textbf{Onl.}} &
\multicolumn{3}{c|}{\textbf{ScanNet (Static)}} &
\multicolumn{3}{c|}{\textbf{Sintel}} &
\multicolumn{3}{c}{\textbf{TUM-dynamics}} \\
\cmidrule(lr){4-6}\cmidrule(lr){7-9}\cmidrule(lr){10-12}
& & &
ATE$\downarrow$ & RPE trans$\downarrow$ & RPE rot$\downarrow$ &
ATE$\downarrow$ & RPE trans$\downarrow$ & RPE rot$\downarrow$ &
ATE$\downarrow$ & RPE trans$\downarrow$ & RPE rot$\downarrow$ \\
\midrule
Robust-CVD~\cite{kopf2021robust-cvd}   & \checkmark &  & 0.227 & 0.064 & 7.374 & 0.360 & 0.154 & 3.443 & 0.153 & 0.026 & 3.528 \\
CasualSAM~\cite{zhang2022casualsam}    & \checkmark &  & 0.158 & 0.034 & 1.618 & \underline{0.141} & \textbf{0.035} & \textbf{0.615} & \underline{0.071} & \textbf{0.010} & 1.712 \\
DUSt3R-GA~\cite{wang2024dust3r}    & \checkmark &  & \underline{0.081} & 0.028 & 0.784 & 0.417 & 0.250 & 5.796 & {0.083} & 0.017 & 3.567 \\
MASt3R-GA~\cite{leroy2024mast3r}    & \checkmark &  & \textbf{0.078} & \underline{0.020} & \textbf{0.475} & 0.185 & 0.060 & 1.496 & \textbf{0.038} & \underline{0.012} & \textbf{0.448} \\
MonST3R-GA~\cite{zhang2024monst3r}  & \checkmark &  & 0.077 & \textbf{0.018} & \underline{0.529} & \textbf{0.111} & \underline{0.044} & \underline{0.869} & 0.098 & 0.019 & \underline{0.935} \\
\midrule
Spann3R~\cite{wang2025Spann3r} &  & \checkmark 
& \underline{0.096} & \underline{0.023} & {0.661} 
& 0.329 & 0.110 & 4.471 
& 0.056 & 0.021 & 0.591 \\

CUT3R~\cite{wang2025cut3r} &  & \checkmark 
& 0.099 & \textbf{0.022} & \underline{0.600} 
& \textbf{0.213} & \underline{0.066} & \underline{0.621} 
& \textbf{0.046} & \underline{0.015} & \underline{0.473} \\

Point3R~\cite{wu2025point3r} &  & \checkmark 
& 0.097 & 0.035 & 2.791 
& 0.442 & 0.154 & 1.897 
& 0.058 & 0.031 & 0.758 \\

Ours &  & \checkmark 
& \textbf{0.086} & 0.026 & \textbf{0.583} 
& \textbf{0.213} & \textbf{0.059} & \textbf{0.616} 
& \underline{0.049} & \textbf{0.014} & \textbf{0.463} \\
\bottomrule
\end{tabular}
}
\caption{\textbf{Camera Pose Estimation Evaluation}. Best and second-best results are highlighted separately for optimization-based and online methods. Since global optimization-based methods generally have an advantage over online methods in camera pose estimation, we compare the performance of global optimization approaches and online approaches separately.
}
\label{tab:camera_pose_eval}
\end{table*}

\subsection{Reconstruction Results}

Table~\ref{tab:mv_recon_7scenes_nrgbd} Quantitative comparison on the \textit{7-Scenes~\cite{shotton2013-7scenes}} and \textit{NRGBD~\cite{azinovic2022nrgbd}} datasets for dense 3D reconstruction. Reconstruction quality is evaluated using \textit{accuracy (Acc)}, \textit{completeness (Comp)}, and \textit{normal consistency (NC)}. Lower values indicate better performance for Acc and Comp, while higher values indicate better performance for NC. We compare our method with both optimization-based approaches (e.g., DUSt3R-GA, MASt3R-GA, MonST3R-GA) and online memory-based methods (Spann3R, CUT3R, and Point3R). The columns labeled \textit{Optim.} and \textit{Onl.} indicate whether a method relies on global alignment optimization or operates in an online streaming manner. As shown in the table, our method achieves the best or competitive results across most metrics on both datasets, demonstrating improved reconstruction accuracy and completeness.
The discrepancy between strong Acc/Comp and relatively lower NC stems from the fact that our method prioritizes accurate and informative point-level representations through selective caching, while lacking explicit constraints on local surface continuity and cross-view consistency, which are essential for reliable normal estimation.
Unlike merging-based strategies that implicitly smooth local geometry, our retain-or-replace mechanism preserves distinct structural details but may sacrifice local surface continuity, negatively affecting normal consistency.
Additionally, NC is sensitive to point density and neighborhood distribution, and our adaptive memory strategy may introduce non-uniform sampling, further affecting normal estimation.
Figure.~\ref{fig:visualized_recon} shows the qualitative results of sparse reconstruction on datasets  7scenes and NRGBD. Our method outperforms Point3R for both datasets.

\subsection{Depth Estimation}

Table~\ref{tab:depth_benchmarks} presents quantitative comparisons for monocular depth estimation on the NYU-v2 (static)~\cite{silberman2012nyu-v2}, Sintel~\cite{butler2012sintel}, Bonn~\cite{palazzolo2019bonn}, and KITTI~\cite{geiger2013kitti} datasets. We report the absolute relative error (Abs Rel) and the percentage of inlier pixels with depth error $\delta < 1.25$. Lower values indicate better performance for Abs Rel, while higher values indicate better performance for the inlier ratio. We compare our method with several recent pointmap-based and memory-based approaches, including DUSt3R, MASt3R, MonST3R, Spann3R, CUT3R, and Point3R. As shown in the table, our method achieves competitive or superior performance across multiple datasets, demonstrating strong generalization ability for zero-shot depth estimation in both indoor and outdoor environments.

\subsection{Camera Pose Estiamtion}

Table~\ref{tab:camera_pose_eval} reports quantitative results for camera pose estimation on the ScanNet (static)~\cite{dai2017scannet}, Sintel~\cite{butler2012sintel}, and TUM-dynamics~\cite{sturm2012tum} datasets. We evaluate pose accuracy using Absolute Trajectory Error (ATE), Relative Pose Error in translation (RPE trans), and Relative Pose Error in rotation (RPE rot), where lower values indicate better performance. The table compares our method with both optimization-based approaches (e.g., Robust-CVD, CasualSAM, DUSt3R-GA, MASt3R-GA, and MonST3R-GA) and online methods (Spann3R, CUT3R, and Point3R). The columns \textit{Optim.} and \textit{Onl.} indicate whether the method relies on global optimization or operates in an online streaming manner. As shown in the results, our method achieves competitive performance across the evaluated datasets while maintaining a fully online reconstruction pipeline without additional post-processing.

\subsection{Ablation}

\begin{table*}[t]
\centering
\small
\resizebox{0.95\textwidth}{!}{
\begin{tabular}{l|l|cccccccccc}
\toprule
Method & Metric &
breakfast & complete\_ & green\_room & grey\_white & kitchen &
morning\_a & staircase & geometry & whiteroom & avg \\
\midrule
\multirow{3}{*}{Point3R}
& Acc$\downarrow$ (mean)  & 0.059 & 0.068 & 0.069 & 0.026 & 0.177 & 0.022 & 0.087 & 0.034 & 0.143 & 0.076 \\
& Comp$\downarrow$ (mean) & 0.040 & 0.043 & 0.107 & 0.021 & 0.172 & 0.024 & 0.033 & 0.020 & 0.110 & 0.063 \\
& NC$\uparrow$ (mean)     & 0.855 & 0.869 & 0.802 & 0.900 & 0.809 & 0.861 & 0.774 & 0.872 & 0.769 & \textbf{0.835} \\
\midrule
\multirow{3}{*}{Ours (retain)}
& Acc$\downarrow$ (mean)  & 0.050 & 0.060 & 0.058 & 0.172 & 0.105 & 0.028 & 0.083 & 0.023 & 0.042 & \underline{0.069} \\
& Comp$\downarrow$ (mean) & 0.012 & 0.019 & 0.053 & 0.025 & 0.061 & 0.017 & 0.018 & 0.006 & 0.012 & \underline{0.025} \\
& NC$\uparrow$ (mean)     & 0.784 & 0.800 & 0.776 & 0.735 & 0.782 & 0.789 & 0.689 & 0.842 & 0.741 & \underline{0.771} \\
\midrule
\multirow{3}{*}{Ours (replace)}
& Acc$\downarrow$ (mean)  & 0.048 & 0.069 & 0.060 & 0.195 & 0.120 & 0.027 & 0.085 & 0.023 & 0.047 & 0.075 \\
& Comp$\downarrow$ (mean) & 0.011 & 0.025 & 0.049 & 0.043 & 0.059 & 0.015 & 0.018 & 0.006 & 0.011 & 0.026 \\
& NC$\uparrow$ (mean)     & 0.756 & 0.789 & 0.771 & 0.733 & 0.786 & 0.798 & 0.684 & 0.838 & 0.731 & 0.765 \\
\midrule
\multirow{3}{*}{Ours (random)}
& Acc$\downarrow$ (mean)  & 0.039 & 0.064 & 0.057 & 0.146 & 0.080 & 0.032 & 0.078 & 0.023 & 0.030 & \textbf{0.061} \\
& Comp$\downarrow$ (mean) & 0.009 & 0.019 & 0.059 & 0.022 & 0.034 & 0.018 & 0.017 & 0.006 & 0.009 & \textbf{0.022} \\
& NC$\uparrow$ (mean)     & 0.771 & 0.789 & 0.774 & 0.746 & 0.794 & 0.788 & 0.694 & 0.841 & 0.738 & \underline{0.771} \\
\bottomrule
\end{tabular}
}
\caption{Ablation study on different memory cache update strategies. Best results are highlighted in \textbf{bold} and second-best results are \underline{underlined}. Our retain-or-replace strategy is denoted as `random' in the table.}
\label{tab:mem_cache_ablation}
\end{table*}

To investigate the influence of different pointer update strategies, we conduct an ablation study under three alternative designs. In the first setting, denoted as case `retain', when a newly added feature point is considered redundant with respect to an existing memory point, the existing point is always preserved, and the new point is discarded. In the second setting, referred to as case `random', when redundancy is detected between a new feature point and an existing memory point, one of the two is randomly selected to be retained; that is, the system either keeps the existing point or replaces it with the newly added point with equal probability. In the third setting, termed `replace', whenever redundancy occurs, the existing memory point is directly replaced by the newly added point. The quantitative results of these three strategies are reported in Table~\ref{tab:mem_cache_ablation}, which allows us to analyze the impact of deterministic retention, stochastic updating, and full replacement on reconstruction quality and memory stability.
The experimental results show that the `merge' strategy used in Point3R performs even worse than the retain strategy. This can be partly attributed to the averaging operation in `merge', which degrades the feature representation. In contrast, our final method is more effective than directly applying retain or replace, as it is more likely to preserve critical information rather than being limited by local redundancy or suffering from information loss that leads to forgetting.

\begin{figure}[t]
\begin{center}
   \includegraphics[width=0.95\linewidth]{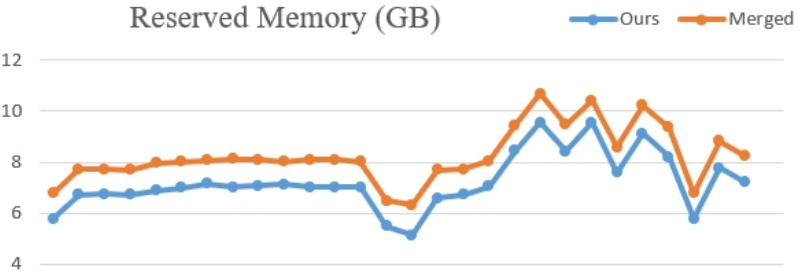}
\end{center}
   \caption{Reserved Memory used by the merged method (Point3R) and our proposed retain-or-replace method for different sequences from two different datasets.}
\label{fig:memory}
\end{figure}

\subsection{Reserved Memory Comparison}
We further evaluate the memory efficiency of our method by comparing the reserved GPU memory against the merged baseline across multiple sequences from different datasets. As illustrated in Figure~\ref{fig:memory}, our approach consistently exhibits lower reserved memory consumption than the merged strategy throughout all evaluated sequences in the Bonn and NRGBD datasets.

Specifically, the merged method exhibits a relatively higher, more fluctuating memory footprint, typically ranging from 7.5 GB to 10.5 GB. In contrast, our method is more stable and generally bounded within a lower range of approximately 6 GB to 9 GB. 
This gap becomes more pronounced in complex scenes, where the merged strategy tends to allocate excessive memory due to redundant feature aggregation and less selective pointer updates.

In contrast, our method effectively controls memory growth by adopting a more discriminative pointer update mechanism, which avoids unnecessary redundancy while preserving critical structural information. As a result, the reserved memory usage is not only reduced but also exhibits improved stability across sequences with varying scene complexity. These results demonstrate that our method achieves a more favorable trade-off between representation capacity and memory efficiency, making it more suitable for scalable and resource-constrained 3D reconstruction scenarios.

\section{Conclusion}

In this work, we studied the problem of \textit{streaming 3D reconstruction from long image sequences}, where observations must be continuously integrated into a persistent spatial representation. A key challenge in this setting is identifying consistent \textit{physical scene points under changing viewpoints}, since appearance similarity alone cannot reliably determine spatial identity.

To address this challenge, we proposed a \textit{ray-aware pointer memory framework} that explicitly incorporates viewing direction into spatial memory representations. By storing both 3D position and observation ray direction within each pointer, the system can reason jointly about geometric proximity and viewpoint consistency. Building on this representation, we introduced a \textit{ray-guided unified observation} that distinguishes local redundancy, novel geometry, and loop revisits within a unified framework. Furthermore, we proposed an \textit{adaptive retain-or-replace pointer update strategy} that preserves informative geometric landmarks while maintaining bounded memory growth.

Despite these advantages, several limitations remain. First, the current framework assumes relatively accurate pose estimation during streaming integration, and large pose errors may still affect memory updates. Second, the retain-or-replace update strategy relies on simple stochastic selection and does not yet fully exploit the information content of observations.

Experimental results on multiple benchmarks demonstrate that the proposed approach improves reconstruction accuracy, depth estimation performance, and pose stability in streaming scenarios.

\bibliographystyle{ACM-Reference-Format}
\bibliography{sample-base}

\appendix

\end{document}